%% file: main.tex
\documentclass[10pt,twocolumn,letterpaper]{article}

\usepackage{cvpr}      

\input{preamble}

\definecolor{cvprblue}{rgb}{0.21,0.49,0.74}
\usepackage[pagebackref,breaklinks,colorlinks,allcolors=cvprblue]{hyperref}

\def\paperID{*****} 
\def\confName{CVPR}
\def\confYear{2026}

\title{Fuel Gauge: Estimating Chain-of-Thought Length Ahead of Time in \\ Large Multimodal Models}

\author{{Yuedong Yang \quad Xiwen Wei \quad Mustafa Munir \quad Radu Marculescu}\\
Chandra Family Department of Electrical and Computer Engineering\\
The University of Texas at Austin\\
{\small\texttt{\{albertyoung,xiwenwei,mmunir,radum\}@utexas.edu}}
}

\begin{document}
\maketitle
\input{sec/0_abstract}    
\input{sec/1_intro}

\input{sec/6_relatedwork}
\input{sec/2_background}

\input{sec/3_cot_fuel_gauge}

\input{sec/4_applications}
\input{sec/5_experiment}

\input{sec/7_conclusion}

{
    \small
    \bibliographystyle{unsrt}
    \bibliography{main}
}

\input{sec/X_suppl}

\end{document}

%% file: preamble.tex
\usepackage{multirow}
\usepackage{graphicx}
\usepackage{hwemoji}
\usepackage{algorithm}
\usepackage{algorithmic}



%% file: sec/0_abstract.tex
\begin{abstract}
Reasoning Large Multi-modality Models (LMMs) have become the de facto choice for many applications.
However, these models rely on a Chain-of-Thought (CoT) process that is lengthy and unpredictable at runtime, often resulting in inefficient use of computational resources (due to memory fragmentation) and sub-optimal accuracy (due to under- and over-thinking).
We observe empirically that the CoT process follows a very simple form, whose behavior is independent of the specific generated samples. This suggests that the CoT length can be estimated ahead of time based on a hidden parameter representing the amount of ``fuel'' available to support the reasoning process.
Based on this insight, we propose \textbf{Fuel Gauge, the first method which extracts this hidden signal and predicts CoT length ahead of time}. We demonstrate the utility on the Fuel Gauge on two downstream tasks: predictive KV cache allocation, which addresses memory fragmentation in LMM serving systems, and CoT length modulation, which mitigates under-thinking and over-thinking.
Extensive experiments on LMMs across text-only, image-text, and video-text question answering benchmarks demonstrate the effectiveness, generalizability, and practical value of our Fuel Gauge. For example, on the GPQA-Diamond benchmark, our Fuel Gauge achieves less than half the CoT length prediction error compared to the baseline; this translates into a 13.37$\times$ reduction in the memory allocation frequency. 
\end{abstract}

%% file: sec/1_intro.tex
\section{Introduction}
\label{sec:intro}

Reasoning Large Multimodality Models (LMM)~\cite{yang2025qwen3, guo2025deepseek, bai2025intern, hurst2024gpt, openai_gpt5_2025, qwen3vl} achieve the state-of-art performance across multiple tasks on multiple modalities.
The remarkable performance of reasoning LMMs originates from their inherent reasoning capability through Chain-of-Thought (CoT)~\cite{guo2025deepseek, yang2025qwen3}: when given an input prompt, the reasoning LMMs automatically divide the problem into a sequence of smaller steps and then conquer each sub-problem incrementally. This enables complex thinking behaviors such as backtracking where the model actively reviews its own reasoning, identifies and corrects potential mistakes. As a result, CoT empowers the LMMs with ``human-like'' reasoning capability, allowing them to perform multi-step reasoning and planning.

However, the benefits of CoT come with significant costs both in computation efficiency and response quality. To support complex reasoning, the CoT sequences for LMMs are typically long. For example, as shown in Figure~\ref{fig:cot_len_dist}, a CoT can easily reach a length of 28k tokens, whereas the expected answer may contain only around 1k tokens. Moreover, due to the auto-regressive nature of LMM where each token's prediction depends on the previously generated tokens, the final CoT length is unknown \textit{a priori}. This uncertainty over a potentially long CoT generation can negatively impact both computational efficiency and reasoning quality.

From a computational perspective, the unpredictable and lengthy CoT process forces LMM serving frameworks to repeatedly allocate small, contiguous memory blocks to store the key-value (KV) cache, which grow proportionally with the CoT length. Frequent small allocations lead to memory fragmentation~\cite{kwon2023efficient, maoefficient, zhang2025jenga, xu2024vtensor}, scattering free memory areas across the address space. Consequently, even though sufficient free memory exists, new allocations may fail due to the lack of a big enough contiguous space.
From a response quality perspective, since LMMs lack the broader context about the task at hand, they may incorrectly estimate the task’s difficulty. This mismatch can cause either over-thinking~\cite{chen2024not, aggarwal2025optimalthinkingbench, peng2025revisiting} or under-thinking~\cite{wang2025thoughts, aggarwal2025optimalthinkingbench}, as the length of CoT does not align well with the true complexity of the task. As a result, the reasoning and final response may be suboptimal. Since the CoT length is not known in advance, it is impossible to intervene or correct the LMM in advance.

To address these limitations, we propose Fuel Gauge, \textit{the first framework to enable CoT length estimation at runtime}.
The key idea is inspired by the human brain where certain chemicals indicate the level of energy available for thinking~\cite{holst2015sleep} and influence the thinking process itself~\cite{reichert2022adenosine}.
Analogously, we hypothesize that LMMs possess an internal ``fuel level'' signal that correlates with their reasoning process: the fuel level starts high when the model begins CoT reasoning and gradually decreases to zero as reasoning progresses.
We identify and extract this hidden fuel level signal using a tiny neural network consisting only 82k parameters, and estimate the CoT length based on the rate of fuel consumption. Through extensive experiments across multiple models, benchmarks, and modalities, we empirically validate our hypothesis and demonstrate the effectiveness of the Fuel Gauge. Our key contributions are:

\begin{itemize}
    \item \textbf{Mathematical Characterization of CoT Length:} We observe that CoT length follows a Bernoulli process and demonstrate empirically that it can be predicted \textit{a priori}.
    \item \textbf{First CoT Length Estimation Framework:} We identify an internal signal in LMMs that indicates the CoT length before reasoning completes. Based on this signal, we develop the first CoT length predictor, Fuel Gauge, and successfully apply it to two downstream tasks, \textit{i.e.}, predictive KV cache allocation and CoT length modulation; this demonstrates the practical value of Fuel Gauge.
    \item \textbf{Extensive Empirical Validation:} Experiments across diverse models, benchmarks, and modalities validate our assumptions and demonstrate the effectiveness of the Fuel Gauge framework. Our approach yields substantially lower prediction error than baseline methods (\textit{e.g.}, lower than half error on GPQA-Diamond), leading to marked improvements on multiple downstream tasks (\textit{e.g.}, 13.37$\times$ improvement in memory allocation).
\end{itemize}

The paper is organized as follows. Section~\ref{sec:relatedwork} reviews relevant work. Section~\ref{sec:background} introduces the background of LMMs. Section~\ref{sec:method} then describes our method in detail. Next, we present two applications enabled our Fuel Gauge in Section~\ref{sec:applications}. Experimental results are presented in Section~\ref{sec:exp} and, finally, Section~\ref{sec:conclusion} summarizes our main contributions.

%% file: sec/6_relatedwork.tex
\section{Related Work}
\label{sec:relatedwork}

\noindent\textbf{CoT in LMMs:} CoT prompting~\cite{wei2022chain} enables LMMs to perform step-by-step reasoning on complex tasks such as general reasoning~\cite{wang2024mmlu, du2025supergpqa} and mathematical problem solving~\cite{hendrycks2measuring}. By decomposing a problem into intermediate steps, CoT helps LMMs generate structured reasoning trajectories that improve interpretability and accuracy~\cite{chu2024navigate}.
For instance, DeepSeek-R1~\cite{guo2025deepseek} employs GRPO~\cite{shao2024deepseekmath}; Qwen3~\cite{yang2025qwen3} adopts a cold-start followed by reasoning reinforcement learning with GRPO~\cite{shao2024deepseekmath}; and GPT-OSS~\cite{agarwal2025gpt} leverages CoT-based reinforcement learning to improve step-by-step reasoning performance. 

However, despite improved reasoning capability, the length of CoT reasoning remains largely uncontrolled, resulting in inefficiencies such as excessive or insufficient reasoning (over-thinking or under-thinking) and memory fragmentation~\cite{kwon2023efficient}. Our work addresses precisely this gap by studying CoT length predictability and proposing the \textit{Fuel Gauge} predictor to model and regulate the CoT length. 

\noindent\textbf{Steering Vector and Classifier Guidance:} Steering-based control has been explored as a lightweight mechanism to influence generation behavior in LLMs. \cite{subramani2022extracting} introduced steering vectors, \textit{i.e.} directional activations injected into intermediate layers that can steer the model outputs towards desired attributes. This idea has since been adopted for controlled generation~\cite{dathathriplug} and machine unlearning~\cite{li2024wmdp, huu2024effects, shen2025lunar}.
Analogous mechanisms exist in diffusion models. \cite{dhariwal2021diffusion} proposed classifier guidance, which reweights the score function based on class-conditional likelihoods, effectively biasing the sampling trajectory toward desired semantic regions.

Inspired by these ideas, in Section~\ref{sec:method_cot_mod} we introduce the \textit{CoT length modulation}, a new method that guides the CoT process using our Fuel Gauge. Before Fuel Gauge, predicting the CoT length reliably was impossible, making this the first approach to test-time scaling with classifier guidance.

%% file: sec/2_background.tex
\begin{figure}[t!]
    \centering
    \includegraphics[width=0.8\linewidth]{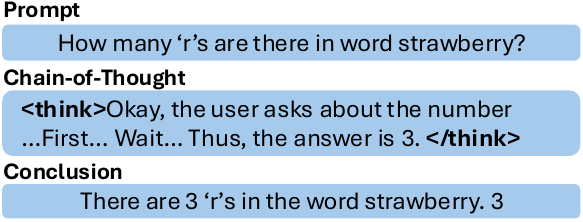}
    \caption{Example of the output of reasoning LMM, which consists of a long \textbf{CoT section} wrapped with special symbols ``\textless think\textgreater'' and ``\textless/think\textgreater'', and a short \textbf{Conclusion} section.}
    \label{fig:cot_example}
\end{figure}

\section{Background on Reasoning LMM}
\label{sec:background}

Reasoning LMMs~\cite{yang2025qwen3, guo2025deepseek, bai2025intern, hurst2024gpt, openai_gpt5_2025} are LMMs fine-tuned with reinforcement learning~\cite{shao2024deepseekmath, schulman2017proximal}, enabling them to perform deep reasoning automatically without any special prompting. As shown in Figure~\ref{fig:cot_example}, the output of the reasoning LMM includes a CoT section wrapped with special tokens ``\textless think\textgreater'' and ``\textless/think\textgreater'' followed by Conclusion.

We model the auto-regressive generation process in CoT as a discrete-time random process with two possible events at each step: $T$ (generation of a terminating token) and $\bar{T}$ (generation of a non-terminating token). A terminating token (\textit{e.g.}, ``\textless/think\textgreater'' in Qwen3~\cite{yang2025qwen3}) ends the CoT, after which the model produces the final answer. Non-terminating tokens correspond to regular words.
Thus, a CoT of length $n$ consists of $(n-1)$ non-terminating events $\bar{T}$ followed by one terminating event $T$.

Let $X_0$ denote the input prompt and $X_{0:(i-1)}$ the concatenation of $X_0$ with generated CoT tokens up to step $(i-1)$. The expected CoT length $N$ can then be expressed as the \textit{first arrival time of $T$} as shown in Equation~(\ref{equ:general_cot_len}):
\begin{equation}
E[N|X_0]=\sum_{n=1}^{\infty}[ n\cdot P(T|X_{0:(n-1)}) \prod_{i=1}^n P(\bar{T}|X_{0:i-1})]
\label{equ:general_cot_len}
\end{equation}
Since each probability depends on the previously generated tokens, $E[N|X_0]$ cannot be computed without sampling the actual CoT trajectories.

%% file: sec/3_cot_fuel_gauge.tex
\begin{figure}
    \centering
    \includegraphics[width=\linewidth]{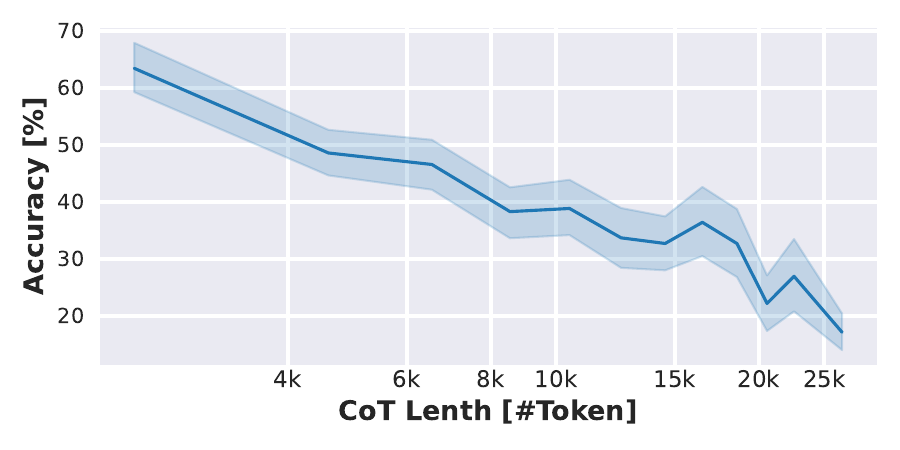}
    \caption{Correlation between Chain-of-Thoughts (CoT) and LMM accuracy collected from Qwen3~\cite{yang2025qwen3}, Qwen3VL~\cite{qwen3vl}, Intern-S1~\cite{bai2025intern} and GLM~\cite{zeng2025glm} across multiple text-only, image-text and video-text benchmarks. Using accuracy as a proxy for task difficulty, we observe a clear negative correlation between CoT length and task difficulty. This trend motivates our hypothesis that CoT length is predictable based solely on the question itself.}
    \label{fig:len_dist}
\end{figure}

\section{Fuel Gauge: First CoT Length Predictor}
\label{sec:method}

\begin{figure*}[t!]
    \centering
    \includegraphics[width=0.95\linewidth]{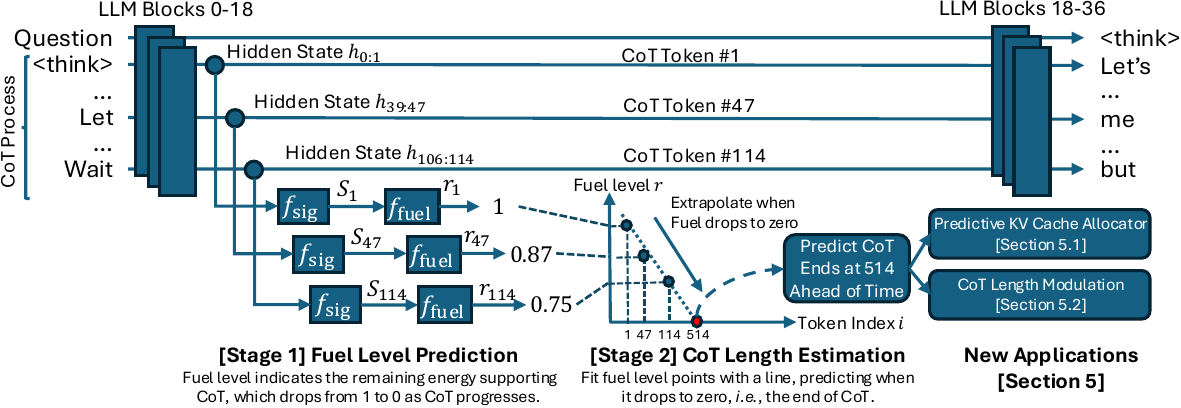}
    \caption{CoT length estimation using the Fuel Gauge. Numbers in the figure are randomly chosen for illustration purpose. See Section~\ref{sec:fuel_gauge_impl} for implementation details. In Stage 1, the hidden signal $S_i$ is extracted using $f_{\text{sig}}$ and the corresponding fuel level is estimated with $f_{\text{fuel}}$. In Stage 2, a linear model is fitted to all predicted fuel-level points, and the zero-crossing point of this line is taken as the final CoT length prediction. Based on this CoT length estimation, we further develop two novel downstream applications (see Section~\ref{sec:applications}).}
    \label{fig:fuel_gauge_proc}
\end{figure*}

In general, the CoT length is hard to predict. However, we show that LMMs possess a unique property that simplifies Equation~(\ref{equ:general_cot_len}), thus enabling CoT length prediction. Based on this observation, we formulate two hypotheses (in Section~\ref{sec:hypo_predictable} and Section~\ref{sec:hypo_fuel}), which provide the foundation for our prediction capability. Section~\ref{sec:len_pred} then describes our CoT length prediction algorithm in detail. Finally, Section~\ref{sec:fuel_gauge_impl} presents the implementation of the Fuel Gauge.

\subsection{Hypothesis I: LMM CoT Length is Predictable}
\label{sec:hypo_predictable}

\noindent\textbf{Observation:} Figure~\ref{fig:len_dist} presents the correlation between CoT length and prediction accuracy collected from multiple LMMs over benchmarks on multiple domains. Considering LMM prediction accuracy as a proxy for task difficulty, we empirically observe a consistent relationship between CoT length and task difficulty across models. Specifically, there is a clear negative correlation between question difficulty (measured by accuracy) and CoT length. 
This finding suggests that CoT length may be predictable directly from the question, without requiring the full CoT sampling process.


Accordingly, we formulate our first hypothesis below:

\noindent\textbf{Hypothesis I:} \textit{The CoT length for reasoning LMM can be predicted before the CoT generation process using a parameter conditioned solely on the input prompt $X_0$.}

To explain, the CoT length is predictable because \textbf{reasoning models implicitly adjust their reasoning depth based on questions difficulty} (like humans). During RLHF, reasoning LMMs are optimized to maximize a reward model which favors long and verbose reasoning traces~\cite{singhal2023long}, while being regularized by KL divergence that discourages deviation from a reference model that produces shorter and more direct responses~\cite{lambert2025reinforcement}. 

The observed CoT length thus reflects a tradeoff between potential reward gains and the cost of KL regularization. More precisely, once an output is considered sufficiently good by the reward model, additional reasoning leads to little marginal reward but higher KL divergence. As a result, unnecessary verbosity is discouraged. Together, these mechanisms encourage reasoning LMMs to produce chains of thought that are short, yet sufficient for solving the task at hand~\cite{wu2025more}. Consequently, CoT length becomes conditioned and thus predictable for a given problem difficulty.

\subsection{Hypothesis II: CoT for LMM is Fuel Powered}
\label{sec:hypo_fuel}

The observation in Figure~\ref{fig:len_dist} motivates us to investigate whether reasoning LMMs contain an internal signal related to the Bernoulli process parameter $p$. Inspired by human cognition, we draw an analogy to how the brain consumes energy during thinking. Neural activity is powered by ATP (adenosine triphosphate) hydrolysis, which releases energy and produces adenosine as a byproduct~\cite{zhu2012quantitative}. As adenosine accumulates and binds to neuronal receptors, it inhibits thinking by reducing alertness and focus~\cite{reichert2022adenosine}. Since adenosine levels reflect ATP consumption, they can be viewed as a proxy for the brain’s energy usage~\cite{holst2015sleep}, \textit{i.e.}, a natural ``fuel gauge'' for cognitive effort.

We hypothesize that reasoning LMMs possess a similar internal signal that indicates their ``fuel'' consumption during reasoning. This signal would start high when reasoning begins, then gradually decrease as thinking progresses, and approach zero when reasoning ends. We refer to this signal as the \textit{Fuel Gauge}. Let $h_{0:i}$ denote hidden states for $X_0$ to $X_i$. Below, we summarize our second hypothesis:

\noindent\textbf{Hypothesis II:} \textit{Assume the CoT consists of $N$ tokens (to be estimated). When generating the $i$-th token, there exists a hidden signal $S_i$ derived from the LMM's hidden states $h_{0:i}$, which can be mapped to a scalar $r_i$ satisfying $r_0=1, r_N=0$ and $r_i>r_j$ for any $i>j$.}

\subsection{CoT Length Prediction Algorithm}
\label{sec:len_pred}

We define $r_i$ as the CoT fuel level, with $f_{\text{sig}}$ mapping hidden states $h_{0:i}$ to the hidden signal $S_i$, and $f_{\text{fuel}}$ mapping hidden signal $S_i$ to fuel level $r_i$.
To estimate CoT length during generation step $i$, we first extract $S_i$ and compute the corresponding fuel level $r_i=f_{\text{fuel}}(S_i)$. We then extrapolate from previous readings $r_0, r_1,\dots,r_i$ to find the zero crossing point $\tilde{N}_i$ where the extrapolated value $\tilde{r}_{\tilde{N}_i}=0$. This $\tilde{N}_i$ serves as the \textit{current estimate} of the CoT length. (Tildes indicate values are obtained through extrapolation.)

To simplify the extrapolation, we use Hypothesis II: \textit{The fuel level $r_i$ decreases \textbf{linearly} with the generation step $i$.} With CoT length $N$ (to be estimated), we fit a linear model:
\begin{equation}
    r_i \approx k \cdot [0\cdots i]^T + 1
\end{equation}
where the slope $k$ is determined from data. We then extrapolate to the point where the fuel level reaches zero ($\tilde{r}_{\tilde N_i}=0$), yielding the CoT length prediction $\tilde{N}_i=-1/k$.

\subsection{Fuel Gauge Implementation}
\label{sec:fuel_gauge_impl}

In this section, we describe the implementation of Fuel Gauge for CoT length prediction, which operates in two stages. As shown in Figure~\ref{fig:fuel_gauge_proc}, at each CoT generation step, Stage 1 extracts and denoises the CoT fuel level, while Stage 2 fits a linear model to estimate the final CoT length. Algorithm~\ref{alg:fuel_gauge} summarizes our implementation.

\noindent\textbf{Stage 1: Fuel Level Estimation} At each time step, Stage 1 first extracts the hidden signal $S_{i}$ using $f_{\text{sig}}$ and then estimates the CoT fuel level using $f_{\text{fuel}}$.
The hidden signal extractor $f_{\text{sig}}$ processes the most recent eight hidden states $h_{i-7:i}$ from a single transformer layer, producing a signal vector $S_i$. The fuel estimator then maps $S_i$ to fuel level $r_i$.

\begin{algorithm}[t]
\caption{Fuel Gauge for CoT Length Prediction}
\label{alg:fuel_gauge}
\begin{algorithmic}[1]
\REQUIRE Hidden states $h_{i-7:i}$; previous fuel level $r_{0:i-1}$; hidden signal extractor $f_{\text{sig}}$; fuel estimator $f_{\text{fuel}}$
\ENSURE Estimated CoT length $\tilde{N}_i$

\STATE \textbf{Stage 1: Fuel Level Estimation}
\STATE $S_i \gets f_{\text{sig}}(h_{i-7:i})$ \COMMENT{Extract hidden signal}
\STATE $r_i \gets f_{\text{fuel}}(S_i)$ \COMMENT{Estimate fuel level}

\STATE \textbf{Stage 2: CoT Length Estimation Update}
\STATE $r_{0:i} \gets \text{Concatenate}(r_{0:i-1}, r_i)$
\STATE Fit linear model $r_{0:i} \approx k \cdot [0\cdots i]^T + 1$
\STATE $\tilde{N}_i \gets -1/k$ \COMMENT{Zero-crossing estimate}
\STATE \textbf{Output:} $\tilde{N}_i$

\end{algorithmic}
\end{algorithm}

To minimize the computational overhead, both networks are intentionally kept simple, \textit{i.e.}, $f_{\text{sig}}$ contains only an 1D depth-wise convolution layer followed by an 1D point-wise convolution layer~\cite{chollet2017xception}, and $f_{\text{fuel}}$ is a two layer MLP. For example, in the Qwen3-4B~\cite{yang2025qwen3} model (36 layers), we take the hidden states from only the last eight time steps (a $8\times 2560$ matrix) at layer 18 and feed them into $f_{\text{sig}}$ followed by $f_{\text{fuel}}$. With only 128 channels in $f_{\text{sig}}$ and $f_{\text{fuel}}$, the added model size and inference latency are negligible compared to Qwen3-4B LMM (see Section~\ref{sec:exp_overhead}).


%

\noindent\textbf{Stage 2: CoT Length Estimation Update} Using the denoised fuel levels $r_{0:i}$ from previous steps, we fit a linear model $r_{0:i}\approx k \cdot [0\cdots i]^T+1$. The zero-crossing point of this line gives the estimated CoT length, \textit{i.e.}, $\tilde{N}=-1/k$. The value $\tilde{N}_i$ serves as the prediction for the total CoT length.

\noindent\textbf{Neural Network Training}
We train $f_{\text{sig}}$ and $f_{\text{fuel}}$ jointly using 200 CoT traces collected from general question answering benchmarks, more precisely the first 200 questions from MMLU~\cite{hendrycks2020measuring} for text-only reasoning model and MMMU~\cite{yue2024mmmu} for text–vision reasoning model. During CoT generation, we record the hidden states $h_i$ from the chosen layer. During training, we randomly sample an 8-step CoT hidden state segment $h_{i-8:i}$ and train the network to predict $r_{i}=f_{\text{fuel}}(f_{\text{sig}}(h_{i-7:i};\theta_{\text{sig}});\theta_{\text{fuel}})$, where $\theta_{\text{sig}}$ and $\theta_{\text{fuel}}$ are trainable parameters. The target value is the normalized token index $(1-\frac{i}{N})$ and objective minimizes the smooth L1 loss~\cite{girshick2015fast} $L_{SL1}$ as shown in Equation~(\ref{equ:fuel_gauge_obj}):
\begin{equation}
\min_{\theta_{\text{sig}}, \theta_{\text{fuel}}} L_{SL1}(f_{\text{fuel}}(f_{\text{sig}}(h_{i-7:i};\theta_{\text{sig}});\theta_{\text{fuel}}), 1-\frac{i}{N}) \\
\label{equ:fuel_gauge_obj}
\end{equation}

%% file: sec/4_applications.tex
\section{Fuel Gauge Applications}
\label{sec:applications}

In this section, we present two applications of our Fuel Gauge, demonstrating its effectiveness, validating our hypotheses, and highlighting its utility in real-world scenarios.

\subsection{Application 1: Predictive KV Cache Allocator}
\label{sec:method_mem}

Dynamic KV cache allocation in LMMs often leads to fragmentation: frequent memory allocation and release create unusable gaps, preventing large contiguous tensor allocation and causing out-of-memory errors even when memory is available~\cite{kwon2023efficient, maoefficient, zhang2025jenga, xu2024vtensor}. By predicting CoT length in advance, we can estimate the memory usage and allocate proactively, thus reducing allocation frequency and fragmentation.

We can allocate memory based on the predicted CoT length at the start of generation. If the allocated memory is insufficient, we can reallocate more memory according to the updated prediction. This repeats until CoT generation completes. The primary evaluation metric is the number of memory allocations: fewer allocations indicate better efficiency and reduced fragmentation.

\subsection{Application 2: CoT Length Modulation}
\label{sec:method_cot_mod}

Inspired by classifier-guided generation~\cite{dhariwal2021diffusion} and steering vectors~\cite{subramani2022extracting, dathathriplug, li2024wmdp, huu2024effects, shen2025lunar}, we use the Fuel Gauge to modulate the CoT length by adjusting hidden states $h_i$ to achieve a desired fuel level $r_{\text{target}}$. Prior to our work, predicting reliably the CoT length is not feasible, making Fuel Gauge the first approach for test-time scaling with classifier guidance.
We formalize this as an optimization problem below:
\begin{equation}
\min_{\Delta h_{i}} J=|f_{\text{fuel}}(f_{\text{sig}}(h_{i-7:i-1}, h_{i}+\Delta h_{i};\theta_{\text{sig}});\theta_{\text{fuel}}) - r_{\text{target}}|
\label{equ:cot_mod_obj}
\end{equation}

Since $f_{\text{fuel}}$ and $f_{\text{sig}}$ are differentiable, a natural approach is gradient ascend/descent: $\Delta h_i=\eta\frac{\partial J}{\partial h_i}$ where $\eta$ is a scalar. However, in practice, since gradient magnitudes vary across models, pinpointing a good set of $\eta$ applicable to all models is difficult. We address this by applying normalized gradient ascend/descend~\cite{zhao2021convergence}, as shown in Equation~(\ref{equ:cot_mod_upd}):
\begin{equation}
h_{i}:=h_{i}+\eta \cdot \frac{\partial J}{\partial h_i} / ||\frac{\partial J}{\partial h_i}||_2
\label{equ:cot_mod_upd}
\end{equation}

We call $\eta$ the \textit{CoT modulation factor}, where positive $\eta$ values increase the fuel level (longer CoT), while negative $\eta$ values decrease the fuel level (shorter CoT).
Section~\ref{sec:exp_cot_mod} demonstrates that changes in $\eta$ reliably influence both CoT length and response quality, validating our method as the first to achieve test-time scaling with classifier guidance and confirming our hypotheses in Sections~\ref{sec:hypo_predictable} and~\ref{sec:hypo_fuel}.

%% file: sec/5_experiment.tex
\section{Experiments}
\label{sec:exp}

In this section, we evaluate our Fuel Gauge and its application. We first evaluate the fuel level prediction accuracy (Stage 1, Section~\ref{sec:fuel_gauge_impl}) in Section~\ref{sec:exp_fuel}, which forms basis of our method and its applications. Next, we test the CoT length prediction accuracy (Stage 2, Section~\ref{sec:fuel_gauge_impl}) in Section~\ref{sec:exp_cot_len}. Finally, in Sections~\ref{sec:exp_mem} and \ref{sec:exp_cot_mod}, we evaluate our Fuel Gauge on two downstream tasks, namely memory allocation prediction and CoT modulation, demonstrating the effectiveness and practical potential of our method. 

Due to space limitations, additional content is provided in the Appendix: Section~\ref{sec_add:exp_setup} describes the experimental setup in detail, Section~\ref{sec_add:exp} presents additional experimental results, Section~\ref{sec_add:ablation} reports the ablation study of our Fuel Gauge, and Section~\ref{sec_add:overhead} evaluates its computational overhead.

\noindent\textbf{Validation of Hypotheses:} We validate the hypotheses proposed in Sections~\ref{sec:hypo_predictable} and~\ref{sec:hypo_fuel} through two complementary approaches. First, we perform direct validation by evaluating the effectiveness of the Fuel Gauge which relies on both hypotheses on fuel level prediction and CoT length prediction tasks.
Second, we perform indirect validation via contradiction by evaluating the Fuel Gauge on the CoT length modulation task. The rationale is as follows: if CoT length is not predictable (Hypothesis I) or if the signal indicating fuel level does not exist (Hypothesis II), then the reading of Fuel Gauge would result from overfitting to noise superficially correlated with CoT length. Consequently, gradient-based alterations in Section~\ref{sec:method_cot_mod} of the Fuel Gauge would only change the fuel level reading itself without producing meaningful changes in the CoT length. Therefore, if we observe consistent and meaningful changes in the CoT length during modulation, this supports the validity of our hypotheses.

\input{tabs/cot_len_table}

\subsection{Experimental Setup and Benchmarks}
\label{sec:exp_setup}

 All models and benchmarks are publicly available. All experiments are conducted on a single NVIDIA A6000 GPU. For text-only question answering, we evaluate Qwen3-4B and Qwen3-8B~\cite{yang2025qwen3} on the GPQA-Diamond~\cite{rein2024gpqa}, AIME24~\cite{aime2024}, and AIME25~\cite{aime2024} benchmarks.
For image–text question answering, we test Qwen3VL-2B and Qwen3VL-4B~\cite{qwen3vl} on MathVision-m~\cite{wang2024measuring}, a subset of MathVision containing 300 questions.

For video–text question answering, we evaluate Qwen3VL-2B and Qwen3VL-4B on LongVideoBench-15 and LongVideoBench-60~\cite{wu2024longvideobench}, corresponding to subsets with 0–15 s and 15–60 s videos, respectively.

For the text-only models (Qwen3-4B and Qwen3-8B), the Fuel Gauge is trained on CoT traces generated from the first 200 questions in the MMLU benchmark.

For the text-vision models (Qwen3VL-2B and Qwen3VL-4B), the Fuel Gauge is trained on CoT traces from the first 200 questions in the MMMU benchmark.
All evaluations are repeated five times with different random seeds. For AIME24 and AIME25, experiments are repeated ten times due to the small benchmark size.

As this work is the first to introduce the fuel level and CoT length predictions, we choose several baseline methods for comparison.

\noindent\textbf{Baselines for Fuel Level Prediction:} 
\begin{itemize}
    \item \textbf{Mean / Median}: Assume all CoTs share the average or median length. The predicted fuel level is the current CoT position divided by this assumed length ($N$).
    \item \textbf{End-of-CoT (EoC) Prob}: Uses the probability of the special End-of-CoT token (\textit{e.g.}, ``\textless/think\textgreater'', shown in Figure \ref{fig:cot_example}) at each CoT step as the predicted fuel level.
\end{itemize}
\noindent\textbf{Baselines for CoT Length Prediction:} 
\begin{itemize}
    \item \textbf{Mean / Median}: Same assumption as above.
    \item \textbf{Direct}: Predicts the CoT length directly using the same inputs and model as the Fuel Gauge (see Section \ref{sec:fuel_gauge_impl}).
\end{itemize}
\noindent\textbf{Baseline for Predictive KV Cache Allocator:} ``\textbf{HF}'' is the default KV Cache allocation strategy adopted by the HuggingFace~\cite{wolf-etal-2020-transformers} framework, which allocates a block of KV cache for 16 tokens when the model runs out of memory.

\noindent\textbf{Baseline for CoT Length Modulation:} We use CoT without modulation as the baseline, \textit{i.e.}, $\eta=0$ in Equation~(\ref{equ:cot_mod_upd}).

Due to the large variance in ground-truth values, we use the \textit{relative mean absolute error} (rMAE) as the evaluation metric for both fuel level and CoT length prediction.

Given $M$ evaluation samples, where each sample $i$ has a CoT of length $N_i$, and at each step $t$ we have a prediction $y_{i,t}$, and ground truth $\hat{y}_{i,t}$, rMAE is defined as:
\begin{equation}
    \text{rMAE}=\frac{\sum_{i=1}^{M}\sum_{t=1}^{N_i}|y_{i,t}-\hat{y}_{i,t}|}{\sum_{i=1}^{M}\sum_{t=1}^{N_i}|\hat{y}_{i,t}|}
\end{equation}

\begin{table}[t]
\resizebox{\columnwidth}{!}{%
\begin{tabular}{lcccc}
\toprule
\multirow{2}{*}{\textbf{Method}} & \multicolumn{2}{c}{\textbf{GPQA-Diamond}} & \multicolumn{2}{c}{\textbf{MathVision-m}} \\
                                 & \textbf{Qwen3-4B}   & \textbf{Qwen3-8B}   & \textbf{Qwen3VL-2B} & \textbf{Qwen3VL-4B} \\
                                 \midrule
Mean                             & 0.2860              & 0.2501              & 0.3399              & 0.3991              \\
Median                             & 0.3161              & 0.2743              & 0.4620              & 0.5611              \\
EoC Prob                         & 0.4999              & 0.4999              & 0.4999              & 0.4999              \\
\textbf{Fuel Gauge}              & \textbf{0.1588}     & \textbf{0.1322}     & \textbf{0.1186}     & \textbf{0.1321}    
\\\bottomrule
\end{tabular}%
}
\caption{Evaluation of the accuracy of fuel level estimation. All number are relative mean absolute error, so smaller is better.}
\label{tab:evalfuel}
\end{table}

\subsection{Fuel Level Prediction}
\label{sec:exp_fuel}

Table~\ref{tab:evalfuel} presents the results of evaluating fuel level estimation (Section~\ref{sec:fuel_gauge_impl}). On both text-only and text-image benchmarks, \textit{i.e.}, GPQA-Diamond and MathVision-m, our Fuel Gauge significantly outperforms baseline methods, reducing estimation error by half or more. This strong performance provides a solid foundation for all subsequent tasks.

\begin{figure}[t]
    \centering
    \includegraphics[width=0.84\linewidth]{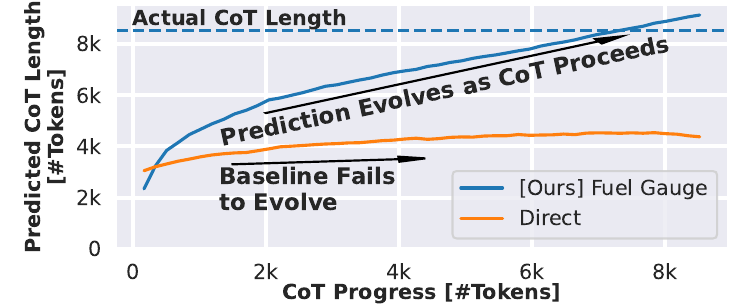}
    \caption{CoT length prediction result with Qwen3-8B model on GPQA-Diamond benchmark. The prediction of our Fuel Gauge evolves as CoT progresses while the baseline fails.}
    \label{fig:cot_len_exp}
\end{figure}

In contrast, static estimations such as “Mean” and “Median” perform poorly, particularly on the MathVision benchmark, where the error reaches 0.5611. This is due to the high variance in CoT lengths. For instance, as shown in Figure~\ref{fig:cot_len_dist}, CoT lengths are widely scattered between 0 and the manually set maximum, so using the mean or median alone cannot adequately capture the distribution.
The “EoC Prob” method performs even worse. Since the probability of the special token marking the end of a CoT is nearly zero most of the time except when the LMM decides to stop, the fuel level estimate rMAE error remains around 0.4999, regardless of the actual CoT lengths across benchmarks.

Overall, these results validate our Fuel Gauge and provide direct support for Hypothesis II in Section~\ref{sec:hypo_fuel}: there exists a signal indicating the fuel level in CoT.

\input{tabs/memory_table}

\subsection{CoT Length Prediction}
\label{sec:exp_cot_len}

\begin{figure*}[t]
    \centering
    \includegraphics[width=0.9\linewidth]{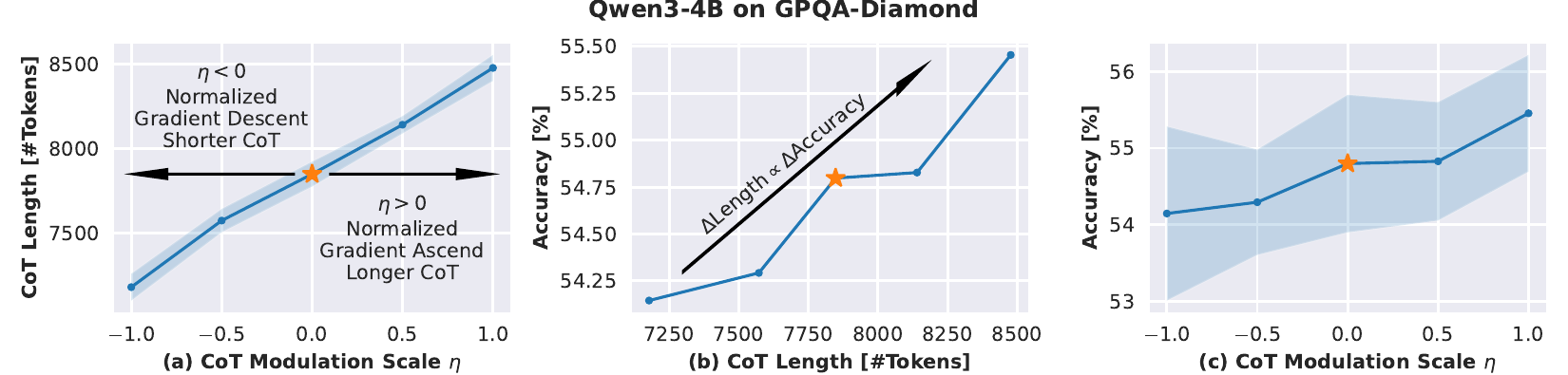}
    \caption{CoT length and LMM accuracy for different CoT modulation factors $\eta$. Results are obtained with Qwen3-4B model on GPQA-Diamond benchmark. Orange star denotes the baseline case where no CoT modulation is applied. Figure (a) shows that Fuel Gauge controls the CoT length linearly. Then Figure (b) shows that the change in CoT length linearly translates to a change in accuracy. Finally Figure (c) shows that based on the linearity in figures (a) and (b), we can achieve our target and control the accuracy linearly with $\eta$.}
    \label{fig:cot_mod_detail}
\end{figure*}

\begin{figure*}[t]
    \centering
    \includegraphics[width=0.97\linewidth]{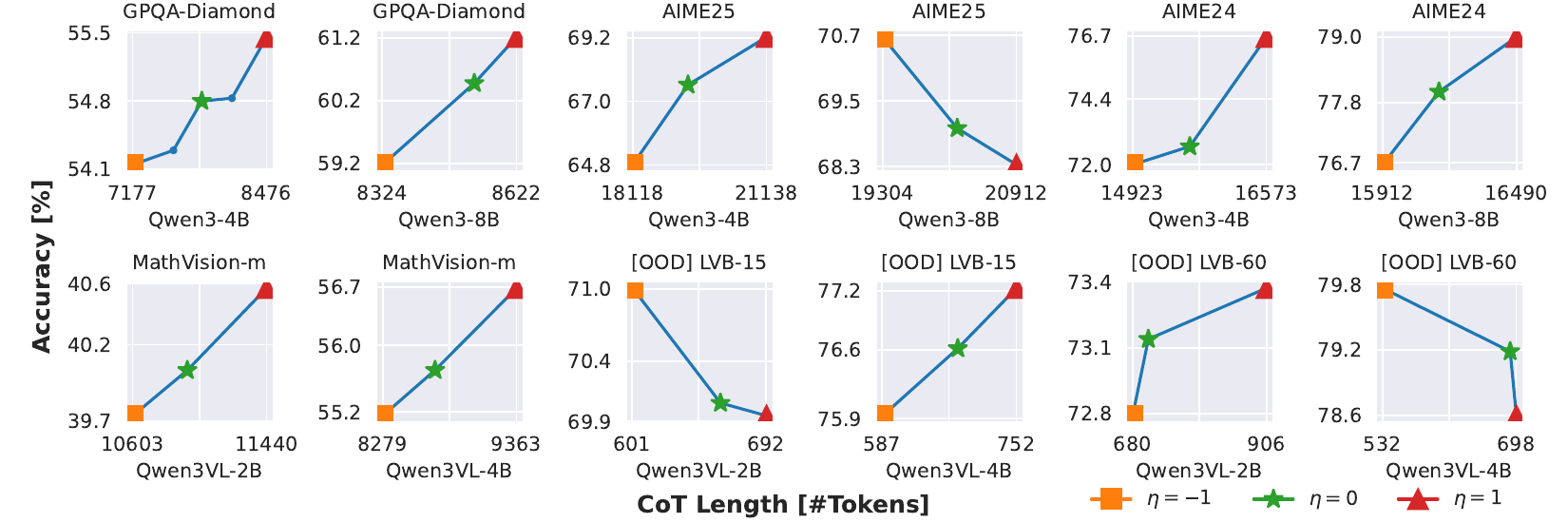}
    \caption{CoT length and LMM accuracy under different CoT modulation factors $\eta$ on more benchmarks and models. ``OOD'' denotes that results are obtained from image-to-video Out-Of-Domain generalization with our Fuel Gauge. ``LVB'' is short for ``LongVideoBench''. By changing $\eta$ from -1 to 1, our Fuel Gauge consistently modulates both CoT length and accuracy linearly, thus demonstrating the wide-applicability of our Fuel Gauge. The green star shows the baseline without modulation. Red and orange markers indicate CoT modulation with positive or negative $\eta$, corresponding to gradient ascent or descent (Equation~\ref{equ:cot_mod_upd}), yielding longer or shorter CoT.}
    \label{fig:cot_mod}
\end{figure*}

Table~\ref{tab:eval_cot_len} presents the experimental results for CoT length prediction. Across all four benchmarks, our Fuel Gauge outperforms all baselines by a large margin. For instance, on GPQA-Diamond with the Qwen3-8B model, our method achieves less than half the error of the “Direct” baseline (see Section~\ref{sec:exp_setup}), which uses the same neural network architecture and data source. Figure~\ref{fig:cot_len_exp} illustrates CoT length prediction at runtime on GPQA-Diamond with Qwen3-8B. As the CoT progresses, the Fuel Gauge effectively incorporates new information, steadily updating predictions toward the ground truth. In contrast, the “Direct” baseline barely adjusts its estimates, even after processing thousands of tokens. This demonstrates not only the effectiveness of our Fuel Gauge, but also directly supports Hypothesis I in Section~\ref{sec:hypo_predictable}: CoT length can be predicted in advance.

\noindent\textbf{Prediction Generalizability:} Our Fuel Gauge exhibits strong generalizability. As described in Section~\ref{sec:exp_setup}, it is trained on only 200 samples from MMLU and MMMU for text-only and multi-modal models, respectively. Consequently, the four benchmarks in Table~\ref{tab:eval_cot_len} present varying levels of generalization challenge.
Specifically, GPQA-Diamond and MathVision-m share the same input modality as MMLU and MMMU, respectively, but differ in task focus, emphasizing deeper knowledge and reasoning. LongVideoBench introduces both modality and task shifts, requiring generalization from image understanding to video reasoning. Despite these variations, our Fuel Gauge consistently provides accurate CoT length estimates.

In contrast, all baseline methods struggle to generalize. For example, on LongVideoBench, the neural network used in the “Direct” baseline, trained on MMMU, fails to produce reasonable CoT lengths due to the large discrepancy between MMMU’s CoT lengths and the much shorter CoTs for LongVideoBench. By dividing CoT length prediction into two stages, \textit{i.e.}, fuel level prediction and CoT length estimation update, our method achieves superior generalizability while using the same model as the baseline.

\subsection{Application 1: Predictive KV Cache Allocator}
\label{sec:exp_mem}

In this section, we evaluate the effectiveness of our Fuel Gauge in the downstream task of predictive KV cache allocation. Table~\ref{tab:mem_alloc} compares the memory allocation requirements of the baseline “HF” method with our Fuel Gauge–based allocator. Across all benchmarks, our method consistently requires significantly fewer memory allocations. For instance, on the MathVision-m benchmark with Qwen3VL-2B, our approach reduces memory allocation frequency by a factor of 9.8.

By leveraging CoT length prediction via Fuel Gauge, we can predict the required KV cache size and allocate it in one shot. This reduces the burden on the memory system by lowering allocation frequency, which in turn minimizes memory fragmentation and improves memory utilization.

\subsection{Application 2: CoT Length Modulation}
\label{sec:exp_cot_mod}

In this section, we evaluate the effectiveness of our Fuel Gauge in modulating the CoT process (Section~\ref{sec:method_cot_mod}). Figures~\ref{fig:cot_mod_detail} and~\ref{fig:cot_mod} present the experimental results for CoT length modulation, with Figure~\ref{fig:cot_mod_detail} showing detailed results for Qwen3-4B on GPQA-Diamond while Figure~\ref{fig:cot_mod} covering broader model–benchmark configurations. From Figure~\ref{fig:cot_mod_detail}, we observe the following:
\begin{itemize}
    \item Figure~\ref{fig:cot_mod_detail}(a): The CoT modulation factor $\eta$ correlates linearly with the CoT Length.
    \item Figure~\ref{fig:cot_mod_detail}(b): Changes in CoT length translate linearly to accuracy.
    \item Figure~\ref{fig:cot_mod_detail}(c): Base on the linearity in previous two findings, accuracy can be linearly controlled by adjusting the CoT modulation factor $\eta$.
\end{itemize}
These findings demonstrate that our Fuel Gauge provides highly effective and stable control over the LMM CoT process: a single factor $\eta$ allows linear adjustment of CoT length, which in turn translates linearly to model accuracy. Figures~\ref{fig:cot_mod} and Table~\ref{tab:cot_lin_corr} further show the broad applicability of the Fuel Gauge across different models and benchmarks, consistently enabling effective linear control of accuracy. This high degree of controllability allows users to efficiently intervene when an LMM is under-thinking or over-thinking.

Moreover, these results provide additional validation for the hypotheses in Sections~\ref{sec:hypo_predictable} and~\ref{sec:hypo_fuel}. If either hypothesis were invalid, the Fuel Gauge would likely fail to achieve stable and consistent linear control over CoT length (see rationale at the beginning of Section~\ref{sec:exp}). The observed linearity therefore supports both hypotheses and confirms the effectiveness of our Fuel Gauge.

\subsection{Ablations and Overhead Analysis}
\label{sec:exp_overhead}

Due to space limitations, we include the ablation study and the overhead analysis in the Appendix. Specifically, Appendix~\ref{sec_add:ablation} presents experiments that justify our design choices by evaluating the Fuel Gauge under:
\begin{itemize}
\item varying numbers of parameters for $f_{\text{sig}}$ and $f_{\text{fuel}}$,
\item different window sizes for $f_{\text{sig}}$, and
\item alternative loss functions in Equation~\ref{equ:fuel_gauge_obj}.
\end{itemize}
Appendix~\ref{sec_add:overhead} further shows that the computational overhead introduced by the Fuel Gauge is negligible.


\begin{table}
\centering
\resizebox{0.9\columnwidth}{!}{%
\begin{tabular}{lccc}
\toprule
\textbf{Modality} & \textbf{$\eta$ to Length} &  \textbf{Length to Acc} & \textbf{$\eta$ to Acc}  \\
\midrule
Text$\rightarrow$Text        & 0.9936       & 0.9754                  & 0.9722                             \\
Image$\rightarrow$Image      & 0.9932       & 0.9988                  & 0.9866                             \\
Image$\rightarrow$Video      & 0.9908       & 0.9835                  & 0.9537                            
\\ \bottomrule
\end{tabular}%
}
\caption{Absolute Pearson correlation among the CoT modulation coefficient $\eta$, CoT length and CoT accuracy (``Acc''). ``Modality'' lists the training and testing modality of our Fuel Gauge. }
\label{tab:cot_lin_corr}
\end{table}

%% file: tabs/cot_len_table.tex
\begin{table*}[t]
\resizebox{\linewidth}{!}{%
\begin{tabular}{lcccccccc}
\toprule
\multirow{3}{*}{\textbf{Method}} & \multicolumn{4}{c}{\textbf{Same Modality, Task-to-Task Generalization, rMAE $\downarrow$}}                 & \multicolumn{4}{c}{\textbf{Image-to-Video \& Task-to-Task Generalization, rMAE $\downarrow$}}                      \\ \cline{2-9} 
                                 & \multicolumn{2}{c}{\textbf{GPQA-Diamond}} & \multicolumn{2}{c}{\textbf{MathVision-m}} & \multicolumn{2}{c}{\textbf{LongVideoBench-15}} & \multicolumn{2}{c}{\textbf{LongVideoBench-60}} \\
                                 & \textbf{Qwen3-4B}   & \textbf{Qwen3-8B}   & \textbf{Qwen3VL-2B} & \textbf{Qwen3VL-4B} & \textbf{Qwen3VL-2B}    & \textbf{Qwen3VL-4B}   & \textbf{Qwen3VL-2B}    & \textbf{Qwen3VL-4B}   \\
                                \midrule
Mean                             & 0.5721              & 0.5003              & 0.6798              & 0.7982              & 3.159                  & 3.706                 & 7.384                  & 4.887                 \\
Median                           & 0.5967              & 0.5212              & 0.7621              & 0.8931              & 3.557                  & 4.122                 & 7.967                  & 5.410                 \\
Direct                           & 0.4932              & 0.5795              & 0.4748              & 0.4965              & 9.833                  & 4.782                 & 10.397                 & 5.877                 \\
\textbf{Fuel Gauge}              & \textbf{0.3185}     & \textbf{0.2732}     & \textbf{0.2934}     & \textbf{0.3139}     & \textbf{0.4834}        & \textbf{0.4527}       & \textbf{0.5049}        & \textbf{0.4645}      
\\\bottomrule
\end{tabular}%
}
\caption{Results for CoT length prediction. Values are in rMAE, hence the lower the better. The best results are highlighted. Baseline methods are introduced in Section~\ref{sec:exp_setup}. GPQA-Diamond and MathVison show the generalizability of Fuel Gauge from general tasks to specialized tasks, and LongVideoBench shows the both the cross-task and cross-modality (\textit{i.e.}, text-image to text-video) generalizability.}
\label{tab:eval_cot_len}
\end{table*}

%% file: tabs/memory_table.tex
\begin{table*}
\resizebox{\linewidth}{!}{%
\begin{tabular}{lcccccccc}
\toprule
\multirow{3}{*}{\textbf{Method}} & \multicolumn{4}{c}{\textbf{Same Domain, Task-to-Task Generalization, \#Allocs $\downarrow$}}                   & \multicolumn{4}{c}{\textbf{Image-to-Video \& Task-to-Task Generalization, \#Allocs $\downarrow$}}              \\ \cline{2-9} 
                                 & \multicolumn{2}{c}{\textbf{GPQA-Diamond}}       & \multicolumn{2}{c}{\textbf{MathVision-m}}       & \multicolumn{2}{c}{\textbf{LongVideoBench-15}}  & \multicolumn{2}{c}{\textbf{LongVideoBench-60}}  \\
                                 & \textbf{Qwen3-4B}      & \textbf{Qwen3-8B}      & \textbf{Qwen3VL-2B}    & \textbf{Qwen3VL-4B}    & \textbf{Qwen3VL-2B}    & \textbf{Qwen3VL-4B}    & \textbf{Qwen3VL-2B}    & \textbf{Qwen3VL-4B}    \\
                                 \midrule
HF~\cite{wolf-etal-2020-transformers}                               & 491.0                  & 533.0                  & 682.6                  & 544.0                  & 41.79                  & 42.82                  & 44.63                  & 43.62                  \\
\textbf{Fuel Gauge}              & \textbf{49.24}         & \textbf{39.87}         & \textbf{69.43}         & \textbf{59.10}         & \textbf{18.56}         & \textbf{23.10}         & \textbf{23.60}         & \textbf{27.81}         \\
\midrule
\textbf{Reduction}               & \textbf{9.971$\times$} & \textbf{13.37$\times$} & \textbf{9.831$\times$} & \textbf{9.205$\times$} & \textbf{2.252$\times$} & \textbf{1.854$\times$} & \textbf{1.891$\times$} & \textbf{1.569$\times$}
\\ \bottomrule
\end{tabular}%
}
\caption{Results for predictive memory allocation. ``\#Allocs'' is short for the number of required memory allocations. Lower is better.}
\label{tab:mem_alloc}
\end{table*}

%% file: sec/7_conclusion.tex
\section{Conclusion}
\label{sec:conclusion}

In this paper, we have presented Fuel Gauge, the first framework to enable CoT length prediction for reasoning LMMs.
To this end, we have addressed challenges arising from CoTs of unknown lengths, including low resource utilization due to memory fragmentation and sub-optimal performance caused by over-thinking or under-thinking.

Observing that CoTs follow a Bernoulli process, we first hypothesize that CoT length is predictable \textit{a priori}. Inspired by the human brain, where certain chemicals provide “fuel” for thinking, we further hypothesize the existence of a hidden “fuel level” signal that reflects the energy available for CoT and can be used to predict its length.

Building on these hypotheses, we have designed the Fuel Gauge, \textit{i.e.}, a compact neural network that can successfully extract this hidden fuel level signal. Using the Fuel Gauge, we can predict CoT length by extrapolating the time step at which the fuel level reaches zero.

Finally, we have shown two practical applications of the Fuel Gauge: predictive KV cache allocation and CoT length modulation. Extensive evaluations across multiple models, benchmarks, and modalities validate our hypotheses and demonstrate the practical value of the Fuel Gauge.

%% file: sec/X_suppl.tex
\appendix
\clearpage
\setcounter{page}{1}
\maketitlesupplementary
\renewcommand{\thetable}{S\arabic{table}}
\renewcommand{\thefigure}{S\arabic{figure}}

\section{More Details on the Experimental Setup}
\label{sec_add:exp_setup}

We perform all experiments on a dedicated server equipped with an AMD EPYC 9554 CPU and eight NVIDIA RTX A6000 GPUs, each with 48 GB of VRAM. Although the machine hosts multiple GPUs, all training and inference experiments for this paper are executed using a single GPU, ensuring reproducibility and eliminating cross-GPU variability. The system runs a standard Linux environment with CUDA-compatible drivers suitable for large-scale model training.

Our implementation uses PyTorch 2.8.0~\cite{pytorch} as the primary deep learning framework. For text-only experiments, we rely on HuggingFace Transformers v4.53.2~\cite{wolf-etal-2020-transformers}; this provides stable support for the language models evaluated in these tasks. For multimodal and other non–text-only settings, we use Transformers v4.57.0, as this version includes updated support for the vision-language models and additional architectural features required by our experiments.

\noindent\textbf{Training Procedure for $f_{\text{fuel}}$ and $f_{\text{sig}}$}
Section \ref{sec:fuel_gauge_impl} provides a high-level overview of the training strategy for the fuel estimator $f_{\text{fuel}}$ and the signal extractor $f_{\text{sig}}$. Here, we elaborate on the full training configuration. Both modules are trained independently using the AdamW optimizer~\cite{adamw} with a base learning rate of $10^{-3}$, and a weight decay coefficient of $10^{-4}$. To ensure stable optimization during the early phase of training, we employ a cosine learning-rate schedule with 1000 warm-up steps~\cite{cosine}, allowing the model to gradually adapt before entering the main decay phase.

We train each model for a single epoch, which is sufficient due to the dense supervision signal and the relatively low-dimensional nature of both predictors. During training, batches are shuffled to improve statistical efficiency, and gradient updates are applied without gradient accumulation on our single-GPU setup. After the training epoch completes, we evaluate the models on a held-out validation set and save the resulting checkpoints for all subsequent inference-time experiments.

\section{Additional Experimental Results}
\label{sec_add:exp}

\subsection{Experimental Results on LMM Intern-S1}

\begin{figure*}[t]
    \centering
    \includegraphics[width=0.95\linewidth]{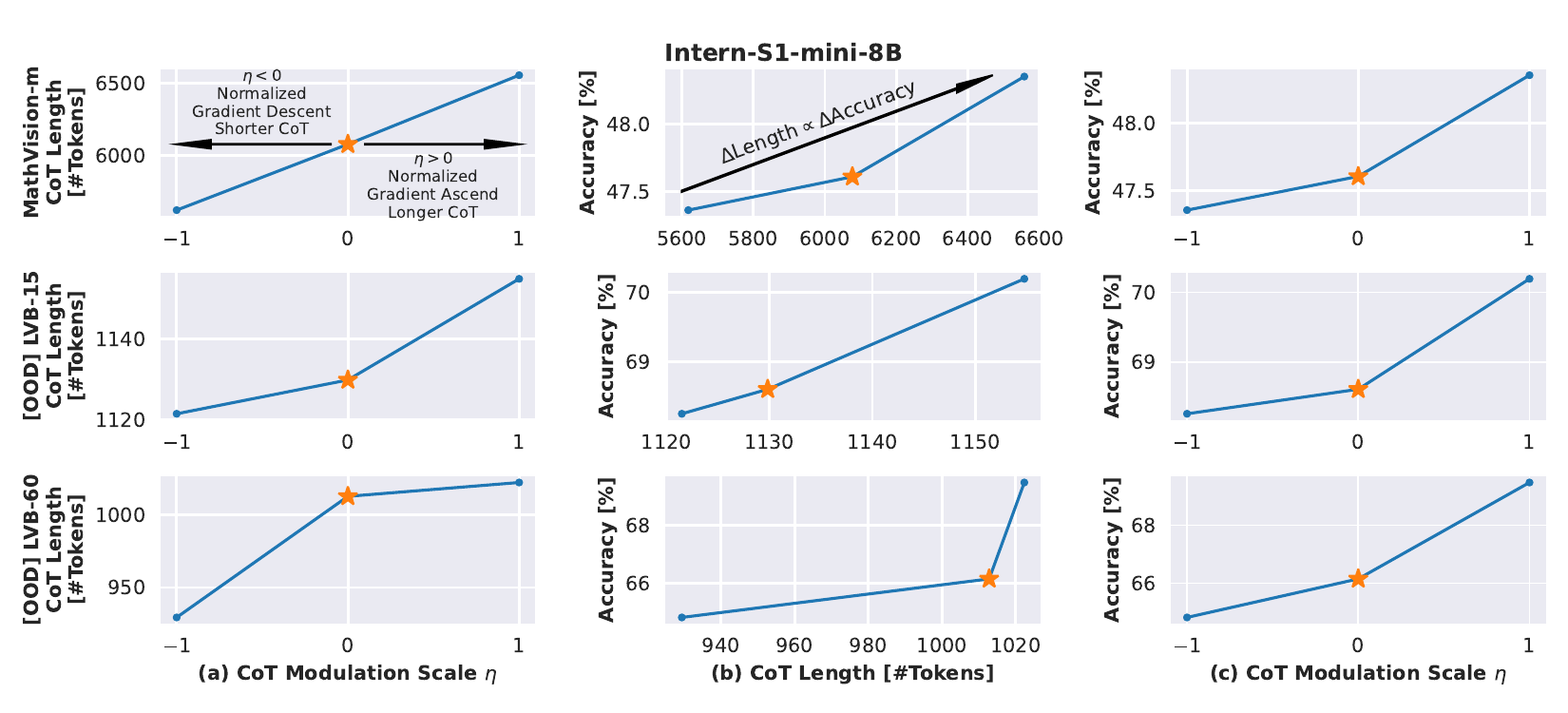}
    \caption{CoT length and LMM accuracy for Intern-S1-mini-8B with different CoT modulation factors $\eta$.  Orange star denotes the baseline case where no CoT modulation is applied. Figure (a) shows that Fuel Gauge controls the CoT length linearly. Figure (b) shows that the change in CoT length quasi-linearly translates to a change in accuracy. Finally, Figure (c) shows that based on the linearity in figures (a) and (b), we can achieve our target and control the accuracy quasi-linearly with $\eta$.}
    \label{fig:exp_intern}
\end{figure*}

\begin{table}[ht]
\centering
\begin{tabular}{lccc}
\toprule
\textbf{Method}     & \textbf{MathVision-m} & \textbf{LVB-15} & \textbf{LVB-60} \\
\midrule
\multicolumn{4}{c}{\textbf{Fuel Level Estimation, rMAE$\downarrow$}}                                                                                                                      \\
\midrule
Mean (Section~\ref{sec:exp_setup})                & 0.8098                & 1.4544                                                               & 1.417                                                                \\
\textbf{Fuel Gauge} & \textbf{0.1311}       & \textbf{0.2583}                                                      & \textbf{0.2831}                                                      \\
\midrule
\multicolumn{4}{c}{\textbf{CoT Length Prediction, rMAE $\downarrow$}}                                                                                                                     \\
\midrule
Mean (Section~\ref{sec:exp_setup})               & 0.8550                & 0.8598                                                               & 1.130                                                                \\
\textbf{Fuel Gauge} & \textbf{0.4194}       & \textbf{0.4902}                                                      & \textbf{0.4525}                                                      \\
\midrule
\multicolumn{4}{c}{\textbf{KV Cache Allocatior, \#Allocations $\downarrow$}}                                                                                                              \\
\midrule
HF~\cite{wolf-etal-2020-transformers}                  & 380.3                 & 71.07                                                                & 63.79                                                                \\
\textbf{Fuel Gauge} & \textbf{15.43}        & \textbf{34.12}                                                       & \textbf{33.06}                                                       \\
\midrule
\textbf{Reduction}  & \textbf{24.6$\times$} & \textbf{2.08$\times$}                                                & \textbf{1.93$\times$}                                               
\\ \bottomrule
\end{tabular}%
\caption{Evaluation results for Intern-S1-mini-8B are reported on MathVision-m, LongVideoBench-15 (LVB-15), and LongVideoBench-60 (LVB-60). MathVision-m demonstrates the generalizability of Fuel Gauge from general tasks to specialized tasks, while LongVideoBench highlights its generalizability across both tasks and modalities (\textit{i.e.}, from text-image to text-video). Across all three benchmarks, Fuel Gauge significantly outperforms the baseline methods.}
\label{tab:exp_intern}
\end{table}

In this section, we expand our evaluation to Intern-S1-mini-8B~\cite{bai2025intern} model. Table~\ref{tab:exp_intern} presents the experimental results on fuel level estimation (Section~\ref{sec:exp_fuel}), CoT length estimation (Section~\ref{sec:exp_cot_len}) and the downstream task predictive KV cache allocator (Section~\ref{sec:exp_mem}). Figure~\ref{fig:exp_intern} shows the results for CoT length modulation (Section~\ref{sec:exp_cot_mod}). Accross all three tasks shown in the Table~\ref{tab:exp_intern}, our Fuel Gauge achieves significantly better results than baseline methods. Moreover, as shown in Figure~\ref{fig:exp_intern}, our Fuel Gauge provides highly effective and stable control over the LMM CoT process: a single factor $\eta$ allows linear adjustment of CoT length, which in turn translates linearly to model accuracy. In summary, these additional results on the Intern-S1-mini-8B model further demonstarate the wide-applicability of our method.

\subsection{Example of Predictive KV Cache Allocator}

\begin{figure*}
    \centering
    \includegraphics[width=\linewidth]{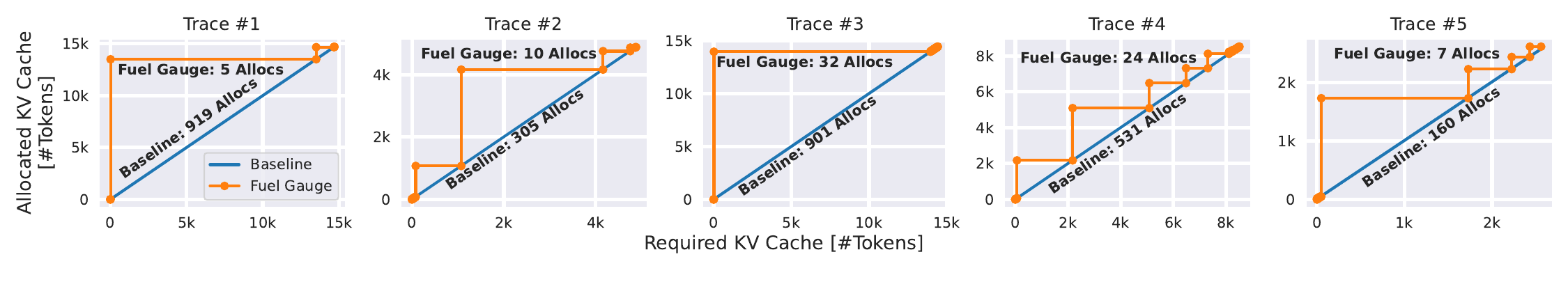}
    \caption{Examples of predictive KV-cache allocation. The baseline allocator (HF, Section~\ref{sec:exp_setup}) performs frequent small, on-demand allocations (diagonal line in plots). With our Fuel Gauge, memory needs are predicted and allocated in larger chunks (stair line in plots), reducing allocation frequency and mitigating fragmentation.}
    \label{fig:exp_mem}
\end{figure*}

Figure \ref{fig:exp_mem} presents five examples of predictive KV-cache allocation (Section \ref{sec:method_mem}). A standard KV-cache allocator cannot anticipate future memory needs and therefore allocates memory only when necessary, resulting in many small allocations—visualized as a diagonal line in Figure \ref{fig:exp_mem}. In contrast, our Fuel Gauge allows the allocator to predict memory requirements in advance, enabling it to allocate a large block of memory in a single step, shown as the ``stairs'' pattern in Figure \ref{fig:exp_mem}. As demonstrated in Table \ref{tab:mem_alloc} and Figure \ref{fig:exp_mem}, this approach greatly reduces the frequency of memory allocation and release, thereby mitigating memory fragmentation.

\subsection{Detailed Results for CoT Length Modulation in Figure~\ref{fig:cot_mod}}

\begin{figure*}
    \centering
    \includegraphics[width=\linewidth]{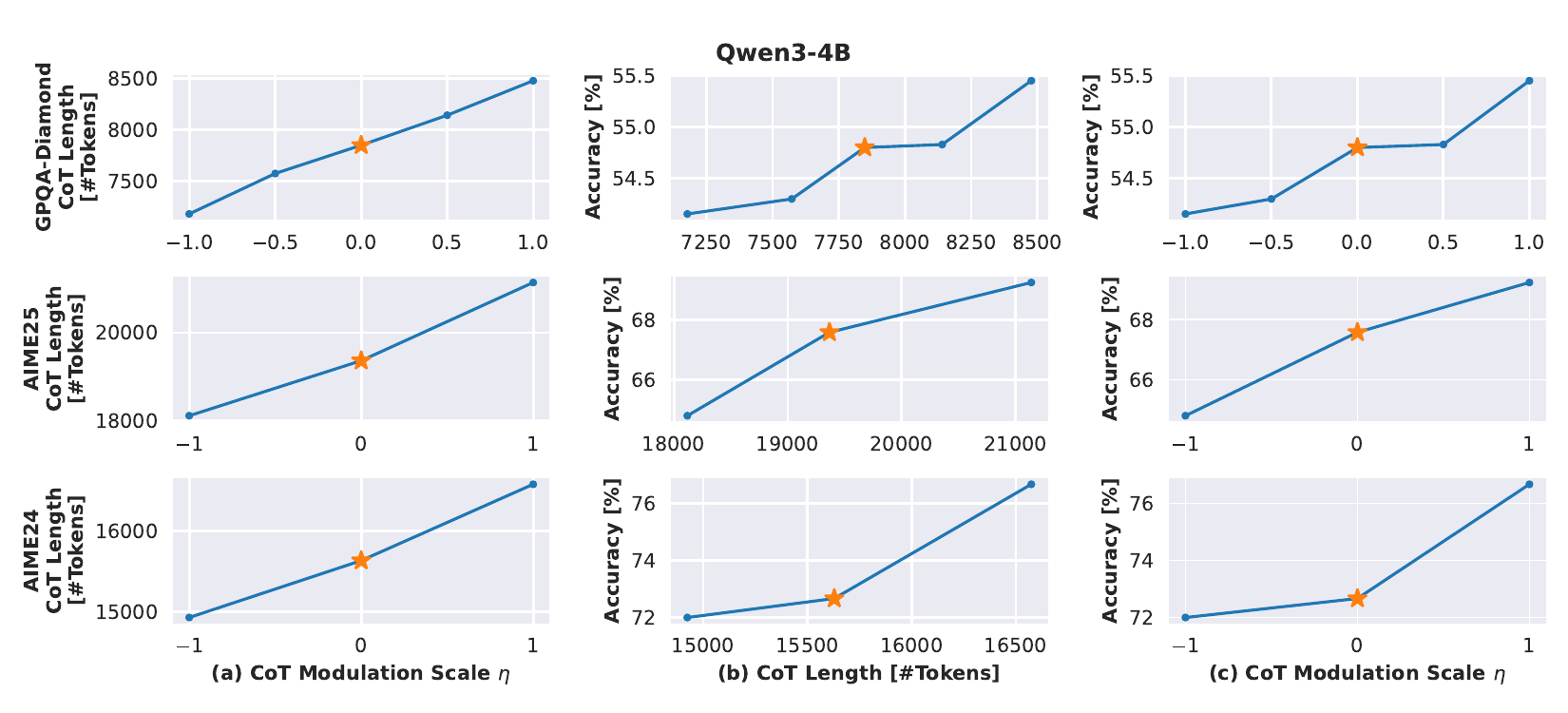}
    \caption{CoT length and LMM accuracy for different CoT modulation factors $\eta$. Results are obtained with Qwen3-4B model on multiple benchmarks. Orange star denotes the baseline case where no CoT modulation is applied. Figure (a) shows that Fuel Gauge controls the CoT length linearly. Figure (b) shows that the change in CoT length quasi-linearly translates to a change in accuracy. Finally Figure (c) shows that based on the quasi-linearity in figures (a) and (b), we can achieve our target and control the accuracy quasi-linearly with $\eta$.}
    \label{fig:add_mod_qwen3_4b}
\end{figure*}

\begin{figure*}
    \centering
    \includegraphics[width=\linewidth]{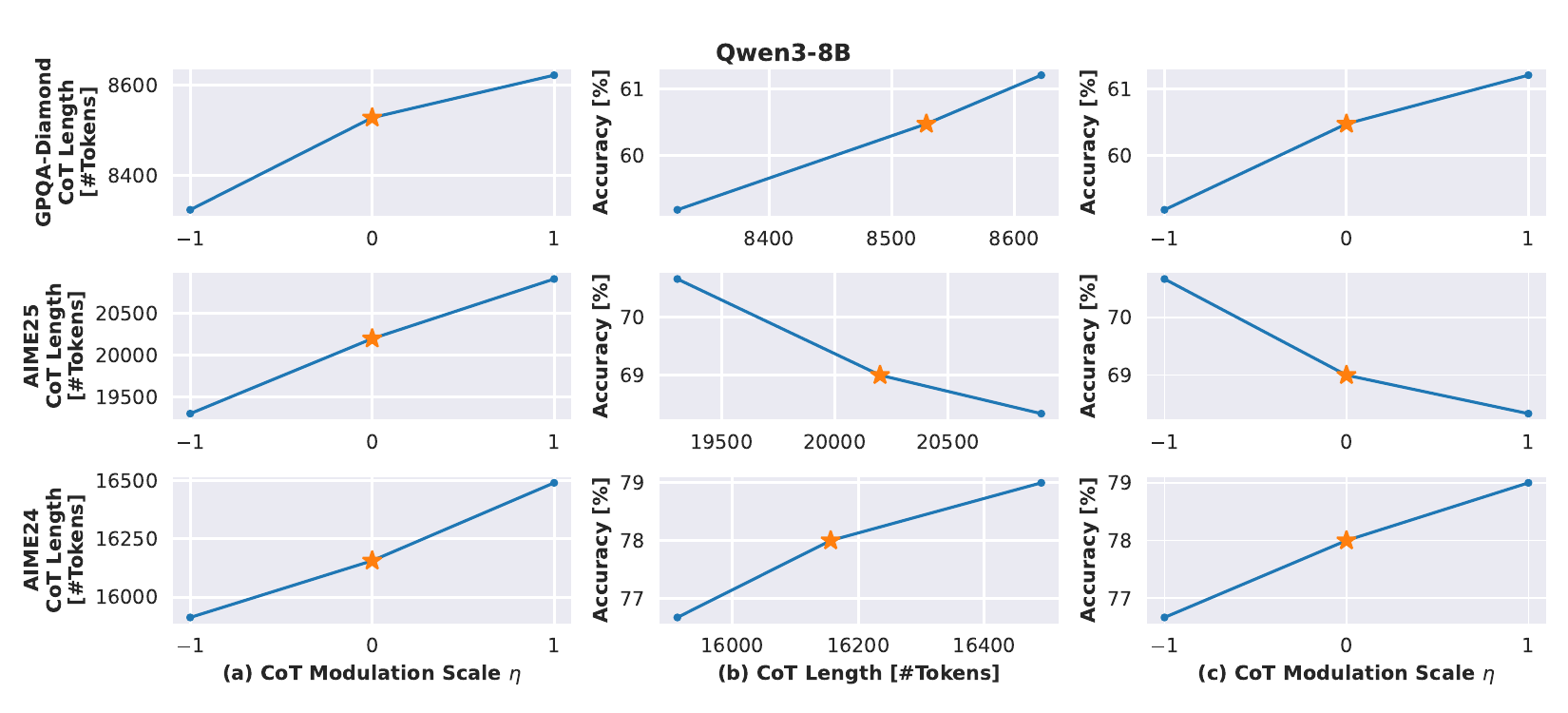}
    \caption{CoT length and LMM accuracy for different CoT modulation factors $\eta$. Results are obtained with Qwen3-8B model on multiple benchmarks. Orange star denotes the baseline case where no CoT modulation is applied. Figure (a) shows that Fuel Gauge controls the CoT length linearly. Figure (b) shows that the change in CoT length quasi-linearly translates to a change in accuracy. Finally Figure (c) shows that based on the quasi-linearity in figures (a) and (b), we can achieve our target and control the accuracy quasi-linearly with $\eta$.}
    \label{fig:add_mod_qwen3_8b}
\end{figure*}

\begin{figure*}
    \centering
    \includegraphics[width=\linewidth]{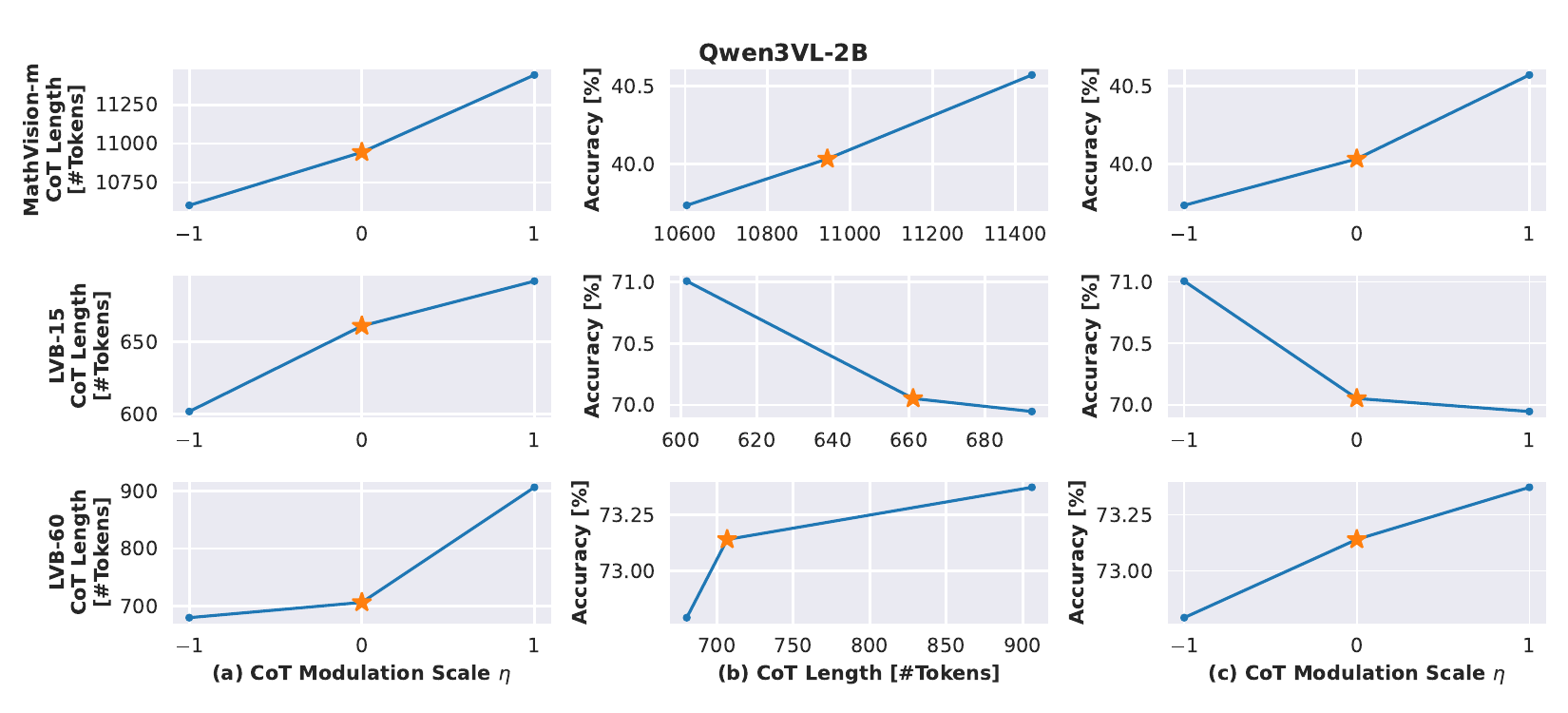}
    \caption{CoT length and LMM accuracy for different CoT modulation factors $\eta$. Results are obtained with Qwen3VL-2B model on multiple benchmarks. Orange star denotes the baseline case where no CoT modulation is applied. Figure (a) shows that Fuel Gauge controls the CoT length linearly. Figure (b) shows that the change in CoT length quasi-linearly translates to a change in accuracy. Finally Figure (c) shows that based on the quasi-linearity in figures (a) and (b), we can achieve our target and control the accuracy quasi-linearly with $\eta$.}
    \label{fig:add_mod_qwen3vl_2b}
\end{figure*}

Figures~\ref{fig:add_mod_qwen3_4b}, \ref{fig:add_mod_qwen3_8b}, \ref{fig:add_mod_qwen3vl_2b}, and \ref{fig:add_mod_qwen3vl_4b} present more detailed results on CoT modulation (placed at the end of the Appendix due to the spacing issues). Across all combinations of LMMs and benchmarks, our Fuel Gauge method consistently enables effective, linear control over CoT length, which in turn leads to correspondingly linear changes in accuracy. This high level of controllability allows users to efficiently intervene when an LMM is either under- or over-thinking.

These results also provide additional support for the hypotheses in Sections \ref{sec:hypo_predictable} and \ref{sec:hypo_fuel}. If either hypothesis were incorrect, the Fuel Gauge would likely fail to maintain stable and consistent linear control of CoT length (see the rationale at the beginning of Section \ref{sec:exp}). The observed linear relationships therefore reinforce both hypotheses and confirm the effectiveness of our Fuel Gauge.

\section{Ablation Study}
\label{sec_add:ablation}

\subsection{Number of Parameters for $f_{\text{sig}}$ and $f_{\text{fuel}}$}
We change the number of parameters for $f_{\text{sig}}$ and $f_{\text{fuel}}$ by varying the number of channels for the convolutional layers and MLP layers of the model.

\begin{figure}[ht]
    \centering
    \includegraphics[width=0.8\linewidth]{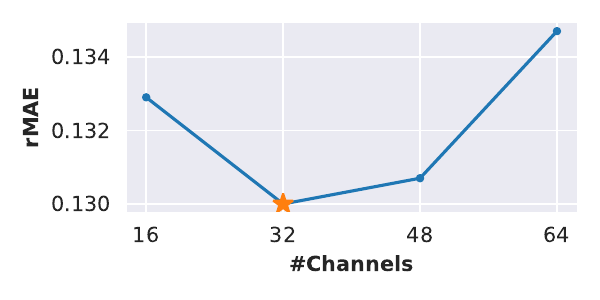}
    \caption{Fuel level estimation accuracy with different number of channels for $f_{\text{sig}}$ and $f_{\text{fuel}}$ for Qwen3-4B on MMLU. Star denotes the adopted design choice for Fuel Gauge.}
    \label{fig:abl_nparam}
\end{figure}

As shown in Figure~\ref{fig:abl_nparam}, our design choice of 32-channels in $f_{\text{fuel}}$ and $f_{\text{sig}}$ yields the best fuel level estimation accuracy. A channel number other than 32 leads to either overfitting or under-fitting issue on the training samples. This supports our design choice for the Fuel Gauge.

\subsection{Window Size for $f_{\text{sig}}$}

\begin{figure}[ht]
    \centering
    \includegraphics[width=0.8\linewidth]{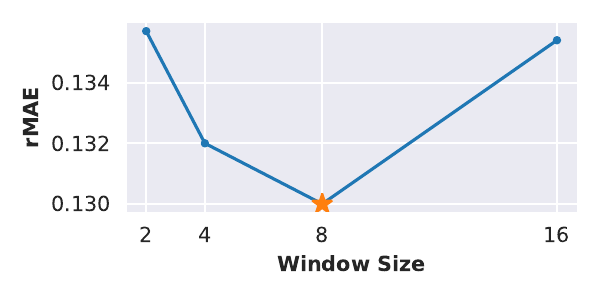}
    \caption{Fuel level estimation accuracy with different sizes of window adopted in $f_{\text{sig}}$ for Qwen3-4B on MMLU. Star denotes the adopted design choice for the Fuel Gauge.}
    \label{fig:abl_windowsize}
\end{figure}

Figure~\ref{fig:abl_windowsize} shows the evaluation result for changing the window size adopted in $f_{\text{sig}}$. For example, as described in Section~\ref{sec:fuel_gauge_impl}, at time step $i$, we use only the most recent 8 hidden states $h_{i-7:i}$ recorded from LMM as the input to $f_{\text{sig}}$, namely, with a window size of 8. As shown in Figure~\ref{fig:abl_windowsize}, changing the window size other than 8 leads to a worse performance. This is because using a larger window size may lead to the staleness in updating prediction while using a smaller window size provides insufficient information to $f_{\text{sig}}$ for hidden signal extraction, whereas our design choice of a window size 8 strikes the balance between agileness in updating the prediction and quality in prediction, thus supporting our design choice.

\subsection{Loss Function in Equation~\ref{equ:fuel_gauge_obj}}

\begin{table}[]
\centering
\begin{tabular}{lc}
\toprule
\textbf{Loss Function}   & \textbf{Test rMAE $\downarrow$} \\
\midrule
MSE                      & 0.1323            \\
MAE                      & 0.1311            \\
Smooth L1 ($\beta=0.1$)   & 0.1317            \\
\textbf{Smooth L1 ($\beta=0.01$)}  & \textbf{0.1300}            \\
Smooth L1 ($\beta=0.001$) & 0.1311           
\\ \bottomrule
\end{tabular}%
\caption{Fuel level estimation accuracy with different loss function in Equation~\ref{equ:fuel_gauge_obj} for Qwen3-4B on MMLU. Bold denotes the adopted design choice for the Fuel Gauge.}
\label{tab:abl_loss_func}
\end{table}

Table~\ref{tab:abl_loss_func} shows the evaluation results for changing the loss functions adopted in the training process for $f_{\text{sig}}$ and $f_{\text{fuel}}$ in Equation~\ref{equ:fuel_gauge_obj}. We consider three loss functions: mean-square error (MSE), mean absolute error (MAE) and smooth L1 loss (smooth L1), where the smooth L1 loss can be formulated as following:
\begin{equation}
    L_{SL1}(y_{i,t}, \hat{y}_{i,t}) = \begin{cases}
        0.5 (y_{i,t} - \hat{y}_{i,t})^2 / \beta, & |y_{i,t} - \hat{y}_{i,t}| < \beta \\
        |y_{i,t} - \hat{y}_{i,t}| - 0.5\beta, & |y_{i,t} - \hat{y}_{i,t}| \ge \beta \\
    \end{cases}
\end{equation}
In Equation~\ref{equ:fuel_gauge_obj}, we adopt smooth L1 with $\beta=0.01$ as our loss function. As shown in Table~\ref{tab:abl_loss_func}, our design choice yields the best performance on the benchmark. This is because the smooth L1 with $\beta=0.01$ strikes the best balance between the MSE loss, which is sensitive to errors but not robust to outliers, and the MAE loss, which is robust to outliers but not sensitive to large errors.

\section{Overhead Analysis}
\label{sec_add:overhead}

Given its small size, the overhead of our Fuel Gauge is negligible. As shown in Table~\ref{tab:overhead}, the Fuel Gauge for the Qwen3-4B model contains only 82k parameters and processes 792.7 tokens per second with a batch size of 1. Increasing the batch size to 32 boosts the throughput to 11k tokens per second. In comparison, the Qwen3-4B model itself has 4 billion parameters with only 22.4 tokens per second throughput. Thus, the additional cost of the Fuel Gauge is indeed minimal.

\begin{table}
\resizebox{\columnwidth}{!}{%
\begin{tabular}{lcc}
\toprule
\textbf{Model}      & \textbf{Throughput} & \textbf{\#Parameters}\\
\midrule
Qwen3-4B                     & 22.4  Token / s   & 4B         \\
Fuel Gauge {[}Batch Size=1{]}  & 792.7  Token / s  &  82.24k       \\
Fuel Gauge {[}Batch Size=32{]} & 11217.3 Token / s  & 82.24k       
\\ \bottomrule
\end{tabular}%
}
\caption{Overhead of Fuel Gauge on NVIDIA A6000 GPU. Results show that the time and memory spent in Fuel Gauge is negligible.}
\label{tab:overhead}
\end{table}

\vspace{10pt}
\noindent\textbf{Figures for CoT Length Modulations in Section~\ref{sec_add:exp} are in Next Page.}

\newpage

\begin{figure*}
    \centering
    \includegraphics[width=\linewidth]{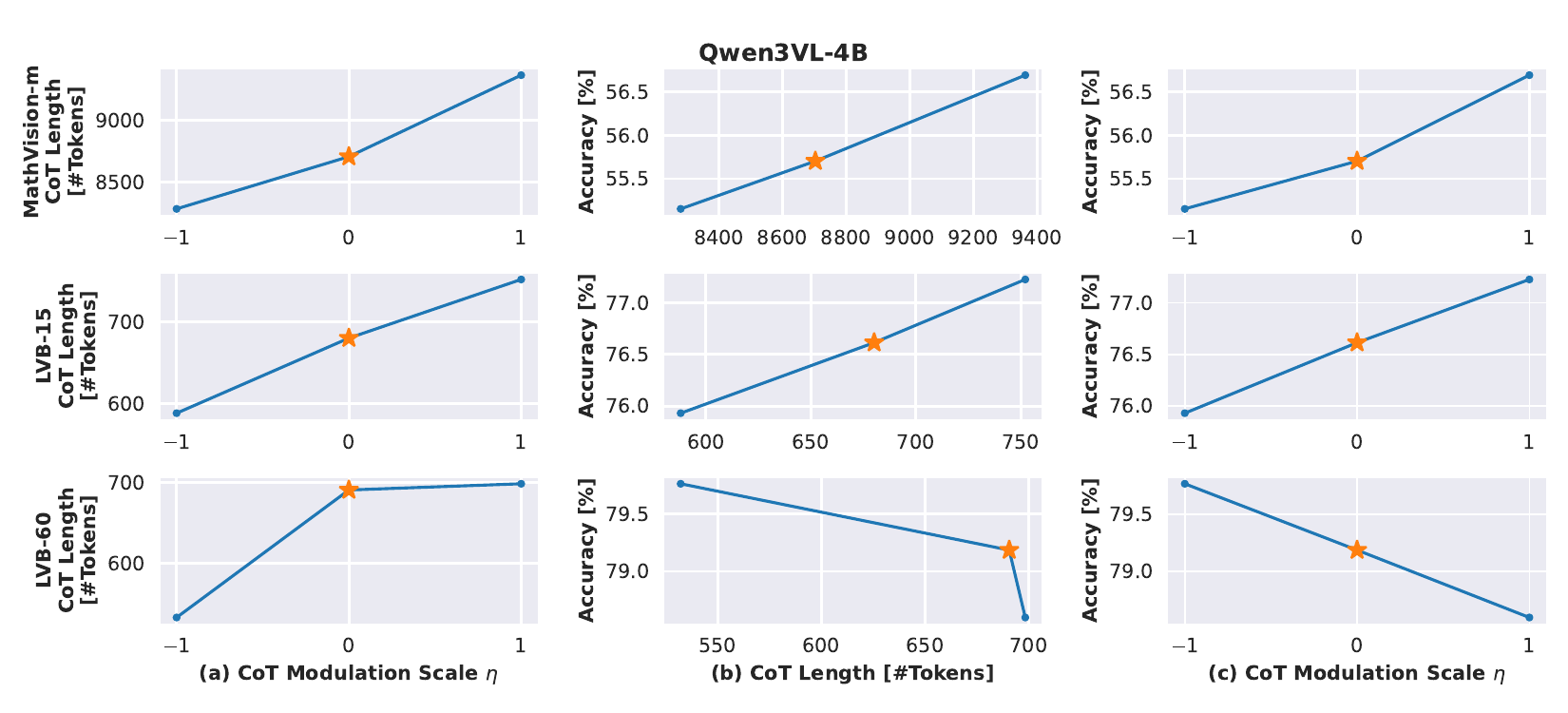}
    \caption{CoT length and LMM accuracy for different CoT modulation factors $\eta$. Results are obtained with Qwen3VL-4B model on multiple benchmarks. Orange star denotes the baseline case where no CoT modulation is applied. Figure (a) shows that Fuel Gauge controls the CoT length linearly. Figure (b) shows that the change in CoT length quasi-linearly translates to a change in accuracy. Finally Figure (c) shows that based on the quasi-linearity in figures (a) and (b), we can achieve our target and control the accuracy quasi-linearly with $\eta$.}
    \label{fig:add_mod_qwen3vl_4b}
\end{figure*}